\pdfoutput=1

\documentclass[11pt]{article}

\usepackage[final]{acl}

\usepackage[utf8]{inputenc}
\usepackage[T1]{fontenc}
\usepackage{hyperref}

\usepackage{times}
\usepackage{latexsym}
\usepackage{microtype}      
\usepackage{xcolor}
\usepackage{graphicx}
\usepackage{caption}
\usepackage{subcaption}
\usepackage{booktabs}
\usepackage{lipsum}  
\usepackage{placeins}
\usepackage{tipa}

\usepackage{tabularx}
\usepackage{multirow}
\usepackage{rotating}

\usepackage{microtype}

\usepackage{inconsolata}
\usepackage{comment}

\usepackage{graphics}
\usepackage{xspace}

\newcommand{\yoruba}{Yor\`ub\'a\xspace}
\newcommand{\ife}{If\d{\`e}\xspace}
\newcommand{\ijebu}{\`Ij\d{\`e}b\'u\xspace}
\newcommand{\ilaje}{\`Il\`aj\d{e}\xspace}

\makeatletter
\renewcommand{\sectionautorefname}{\S\@gobble}
\renewcommand{\sectionautorefname}{\S\@gobble}
\renewcommand{\subsectionautorefname}{\S\@gobble}
\renewcommand{\sectionautorefname}{\S\@gobble}
\renewcommand{\subsectionautorefname}{\S\@gobble}
\makeatother

\newcommand{\menyo}{MENYO-20k\xspace}

\definecolor{purp}{HTML}{791f87}
\definecolor{highlight}{RGB}{255, 255, 0}

\newcommand{\ourdata}{\textsc{Yor\`uLect}}

\title{Voices Unheard: NLP Resources and Models for \yoruba Regional Dialects }

\author{Orevaoghene Ahia\textsuperscript{1,5} \ \ Anuoluwapo Aremu\textsuperscript{4,5} \ \ Diana Abagyan\textsuperscript{1} \ \ Hila Gonen\textsuperscript{1}  \\
\textbf{David Ifeoluwa Adelani\textsuperscript{3,5}} \ \ \ 
\textbf{Daud Abolade\textsuperscript{5}} \ \ \ \textbf{Noah A. Smith\textsuperscript{1,2}} \ \ \ \textbf{Yulia Tsvetkov\textsuperscript{1}} \\
\textsuperscript{1}University of Washington \ \ \textsuperscript{2}Allen Institute for AI \ \
\textsuperscript{3}University College London \\
\textsuperscript{4}Lelapa AI \ \ \textsuperscript{5}Masakhane NLP \\
\href{mailto:oahia@cs.washington.edu}{\texttt{oahia@cs.washington.edu}}
}

\begin{document}
\maketitle
\begin{abstract}
\yoruba---an African language with roughly 47 million speakers---encompasses a continuum with several dialects. Recent efforts to develop NLP technologies for African languages have focused on their standard dialects, resulting in disparities for dialects and varieties for which there are little to no resources or tools. We take steps towards bridging this gap by introducing a new  high-quality parallel text and speech corpus \ourdata{} 
across three domains and four regional \yoruba dialects. To develop this corpus, we engaged native speakers, travelling to communities where these dialects are spoken, to collect text and speech data. Using our newly created corpus, we conducted extensive experiments on (text) machine translation, automatic speech recognition, and speech-to-text translation. Our results reveal substantial performance disparities between standard \yoruba and the other dialects across all tasks. However, we also show that with dialect-adaptive finetuning, we are able to narrow this gap. We believe our dataset and experimental analysis will contribute greatly to developing NLP tools for \yoruba and its dialects, and potentially for other African languages, by improving our understanding of existing challenges and offering a high-quality dataset for further development.
We release \ourdata{} dataset and models publicly under an open license \footnote{Code and data available at \url{https://github.com/orevaahia/yorulect}}.

\end{abstract}

\section{Introduction}
While great strides have been made in developing NLP resources for low-resource languages, the majority of these efforts have been directed towards the ``standard'' dialect of these languages, largely neglecting the long tail of non-standard dialects spoken by millions \cite{faisal2024dialectbench, alam-etal-2024-codet}. Dialects of a language exhibit nuanced yet distinguishable differences in lexicon, pronunciation, spelling, and syntax, mirroring regional, societal, and cultural differences \cite{Chambers_Trudgill_1998}. Usually, a ``standard''
dialect is the dialect with the highest population of speakers, and sometimes the only dialect with a standard orthography \cite{milroy2012authority}. 

African languages are linguistically diverse \cite{adebara-abdul-mageed-2022-towards, siminyu-freshia-2020-ai4d}, yet severely under-resourced. Most of these languages have numerous varieties, (usually regional), some of which are mostly-spoken and lack a standard orthography \cite{batibo2005language, heine2000african}. Developing  language technologies has been incredibly challenging for African languages \cite{nekoto2020participatory, muhammad-etal-2023-afrisenti, ogundepo-etal-2023-cross, adelani-etal-2023-masakhanews, dione-etal-2023-masakhapos, adelani2024irokobench, adelani-etal-2021-masakhaner}, partly due to the scarcity of extensive language resources required for developing systems that are robust to the variations in linguistic features \cite{adebara-abdul-mageed-2022-towards, siminyu-freshia-2020-ai4d}.

To address this problem, in this work we focus on curating dialectal resources for \yoruba, a low-resource language with 47 million native
speakers around the world. \yoruba language is native to Southwestern Nigeria, Republic of Benin, and Republic of Togo. \yoruba encompasses a dialect continuum including several distinct regional dialects \cite{Rowlands_1967}. 
Due to \yoruba's low-resource status, the majority of published NLP work have been done on the Standard \yoruba dialect \cite{ogunremi2024iroyinspeech, aremu2023yorc, ahia-etal-2021-low-resource, dione-etal-2023-masakhapos, shode-etal-2023-nollysenti, ogundepo-etal-2023-cross, akinade-etal-2023-varepsilon, adelani-etal-2023-masakhanews, muhammad-etal-2023-afrisenti, adelani-etal-2021-effect, adebara-etal-2022-linguistically, adebara2021translating, lee-etal-2023-last}.

\begin{table*}[!htbp]
\centering
\resizebox{0.9\linewidth}{!}{

\small
\begin{tabular}{ p{2.5cm} p{2.5cm} p{2.5cm} p{2.5cm} p{2.5cm} p{1cm}} 
\hline
 \textbf{English} & \textbf{Standard} & \textbf{\ijebu} & \textbf{\ife}  &  \textbf{\ilaje} & \textbf{Domain} \\

\hline 

 All the efforts to talk to ASUU chairman failed because he said he has nothing to say & Gbogbo \d{\`o}
 \`i\textcolor{red}{gbi\` y\`anj\'u} l\'ati \textcolor{red}{b\'a al\'aga ASUU} s\d{\`o}r\d{\`o} l\`o j\'as\'i p\`ab\'o nitori \'o ni \`oun k\`o ni ohunk\'ohun l\'ati s\d{o} . 

 & Gbogbo \`i\textcolor{red}{gbi\`y\`anj\'u} l\'ati \textcolor{red}{b\'a al\'aga ASUU} s\d{\`o}r\d{\`o} re jasi afo to ri \'o s\d{o} fo \`o\'un ni ohun k\'ohun l\'ati s\d{o} .
 & Gbogbo \`e\textcolor{red}{gbi\`y\`anj\'u} l\'at\d{e} \textcolor{red}{ba al\'aga ASUU} s\d{\`o}r\d{\`o}  l\d{\`o} j\'as\'i p\`ab\'o tori \'o ghii \`oun n\d{\'e} ihunkihun \'un s\d{o} .
 & Dede \`i\textcolor{red}{gbi\`y\`anj\'u} \'ati \textcolor{red}{b\'a al\'aga ASUU} f\d{\`o} r\`e\'e j\'a ni p\`abo tori \'o f\d{\`o}r\d{\'o} p\'e \'o gh\'un n\'e ir\'u kirun gho f\'e f\`o . & News \\
 \hline

They called unto God in the upper room for the release of the holy spirit .&
W\'on k\'e pe \d{O}l\'orun ni \textcolor{red}{y\`ar\'a} ori \textcolor{red}{\`ok\`e} f\'un i\textcolor{red}{t\'uj\'ade} \textcolor{red}{\d{\`e}mi} mim\'o . & 
W\'on k\'e p\`e \d{O}l\'orun ni \textcolor{red}{y\`ar\'a} ori \textcolor{red}{\`ok\`e} f\'un i\textcolor{red}{t\'u j\'ade} \textcolor{red}{\d{\`e}mi} mim\'o .&
Ig\'an k\'e pe \d{O}l\'oun n\d{\'e} \textcolor{red}{y\`ar\'a} ori \textcolor{red}{\`ok\`e} \'un \d{\`e}\textcolor{red}{t\'uj\'ade} \textcolor{red}{\d{\`e}mi} m\d{\'e}m\d{\'o}
& Gh\'on k\'el\`e kp\`e \d{O}l\'orun ni \textcolor{red}{y\`ar\'a} origho \textcolor{red}{\`ok\`e} gh\'un i\textcolor{red}{t\'uj\'ade} \d{\`e}mi mim\'o . & Religion \\

\hline

 We all look for characteristics that has to do with self-centeredness, and they are similar to this. &
Gbogbo wa la m\'aa \'n w\'a \`aw\d{o}n \`anim\d{\'o} t\'o ni i \d{s}e p\d{\`e}l\'u iwa im\d{o}tara nikan, ìrisi w\d{o}n si j\d{o} \`eyi. &  
Dede wa re n wa iwa \`anim\d{\'o} r\`e nii \d{s}\d{e} p\d{\`e}l\'u iwa im\d{o}lara nikan, irisi w\d{o} si j\d{o} \`iw\'e & 
Gbogbo ria la m\'aa gh\'a in\d{o}n \`an\d{\'e}m\d{\'o} k\d{\'o} n\d{\'e}\d{\'e} i se p\d{\`e}l\'u \`egh\`a \`em\d{o}tara \d{o}ni nik\`an, \d{\`e}risi rian s\`e\`e j\d{o} y\`e\'e .& 
Dede gha r\`e\'e mi f\d{\'e} \`aghan \`anim\'a yii n\d{\'e} i se kp\d{\`e}lu igh\`a im\`otara \d{o}n\d{e} n\`uk\`an, irisi  gh\`an si j\d{o} \`eyi
& Ted Talks\\
 \hline
\end{tabular}}
\caption{Examples of parallel translations across all dialects and domains in \ourdata{}. Words that are unique across all dialects are highlighted in \textcolor{red}{\textit{red}}.}
\label{tab:dialect_examples}
\end{table*}

We introduce the first-ever corpus of high quality, contemporary \yoruba speech and text data parallel across four \yoruba dialects; Standard \yoruba, \ife \textipa{/ i f E /}, \ilaje \textipa{/ i l a dZ E /} and \ijebu\textipa{/ i dZ E b u /} in three domains (religious, news, and Ted talks). This newly curated benchmark, developed with native speakers, can be used in (text-to-text) machine translation (MT), automatic speech recognition (ASR), speech-to-text translation (S2TT), and speech-to-speech translation (STST) tasks. We discuss in detail the data curation process, criteria for data selection, and the steps we took to ensure data quality and integrity  (\autoref{sec:corpus_curation}). We first conduct extensive experiments evaluating the zero-shot performance of recent state-of-the-art models for MT, ASR, and S2TT (\autoref{sec:zero_exps}, \autoref{sec:finetune_exps}). Our results and analysis indicate that current models are not robust enough to handle existing variation in \yoruba dialects. Given these poor results, we proceed to adapt (fine-tune) existing models on our training data across all tasks to boost overall performance. With 802 training instances in each dialect, this approach leads to an average increase of 14 and 5 BLEU points for both MT and S2TT respectively,  as well as a 20-point decrease in word-error-rate for ASR.
Our work aims to motivate the community to build technology for languages alongside their dialects, especially for low-resource dialects of low-resource languages,  as this will promote linguistic diversity, and ensure that technological advancements benefit \emph{all} language communities.

\section{\yoruba and its Regional Dialects}
The \yoruba language is spoken natively by roughly 47 million people in Nigeria\footnote{\url{https://en.wikipedia.org/wiki/Yoruba_language}} and in the neighboring countries of the Republic of Benin and Togo and also Côte d’Ivoire, Sierra Leone, Cuba, and Brazil. In Nigeria, \yoruba speakers are mainly concentrated in the Southwest region, spanning states like Oyo, Ogun, Osun, Ondo, Ekiti, and Lagos, and North Central states like Kogi, and Kwara. 
\begin{figure}[h]
    \centering
    \includegraphics[width=\linewidth]{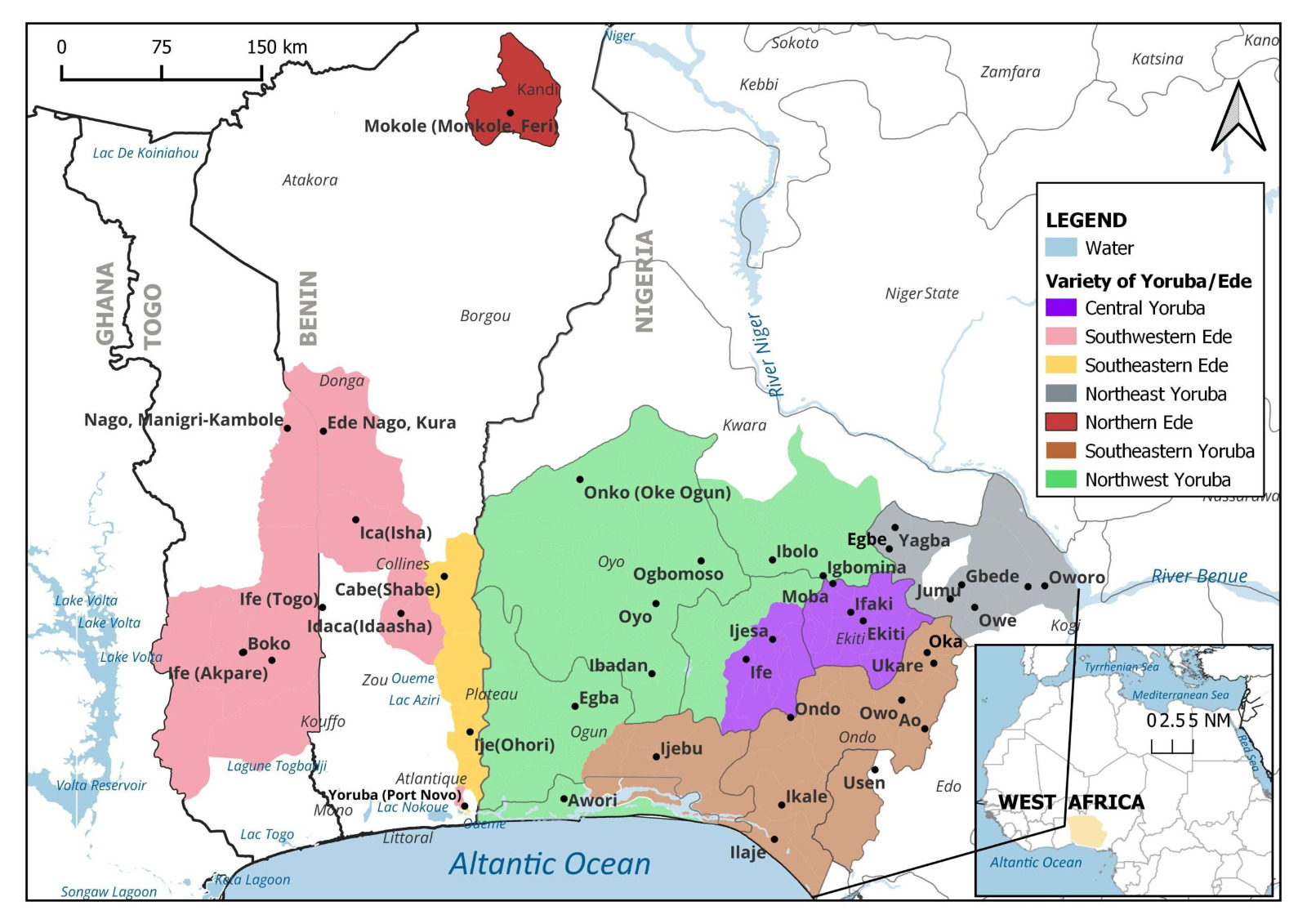}
    \caption{Geographical distribution of \yoruba dialects in West Africa. Map  from \cite{ozburn2023language}.
    }
    \label{fig:regional_map}
\end{figure}

The extensive \yoruba-speaking population and their dispersion across various regions have led to the emergence of geography-specific linguistic variations \cite{ballard1971historical}. 
The number of existing \yoruba dialects is estimated between twelve to twenty-six \cite{ojo1977english, adetugbo1982towards, oyelaran1971yoruba, oyelaran1991africanisms} and the differences present in these dialects are evident in pronunciation, grammatical structure, and vocabulary \cite{adetugbo1982towards, przezdziecki2005vowel, olumuyiwa2009high, arokoyo2019lexicostatistics, olanrewaju2022contrastive}. Also categorized as a 
Volta-Niger language within the Yoruboid subgroup of the Niger-Congo family, \yoruba
is a tonal language with three basic tones: low, middle, and high~\cite{courtenay1969generative, oyetade1988issues}, as well as two or three contour tones.\footnote{A contour tone combines two or more basic tones such as a falling tone made up of a high tone and a low tone, or a rising tone consisting of a low tone followed by a high tone.} Previous research \cite{Adeniyi_2021} has indicated that the phonetic nuances of contour tones are a major distinguishing feature among \yoruba dialects.
 \yoruba dialectal forms in Nigeria can be classified into five regional groupings: Northwest \yoruba (NWY), Northeastern \yoruba (NEY), Central \yoruba (CY), Southwest \yoruba (SWY), and Southeast \yoruba (SEY).
Phonological, lexical, and grammatical differences distinguish these groupings, given the diverse levels of mutual intelligibility among the ``regional'' dialects within each category \cite{arokoyo2019lexicostatistics, olumuyiwa2016vowel, abiodundiachronic}. In this work, our focus lies on \ife, a dialect in the Central Yoruba classification, \ijebu, and \ilaje dialects, which belong to the Southeast Yoruba classification. We display the geographical distribution of \yoruba dialects in West Africa in \autoref{fig:regional_map}. 

\paragraph{Comparative dialectal analysis}
Standard \yoruba, \ife, \ijebu and \ilaje dialects exhibit both similarities and differences in their orthographic representations, morphology, and semantics. For instance, standard \yoruba dialect has fused velar fricative /\textipa{G}/ and labialised voiced velar /\textipa{g\textsuperscript{w}}/ into /w/ \cite{adetugbo1982towards} and our curated data revealed a similar pattern for \ijebu. In contrast, \ife uses /\textipa{G}/ in certain occurrences while \ilaje has heavily retained the /\textipa{g\textsuperscript{w}}/ and /\textipa{G}/ in its representations. 
As a result, at the word level, ``\`aw\d{o}n''
 (3p pl.) is represented similarly in standard dialect and \ijebu but as ``igh\d{o}n'' in \ife and
``\`aghan'' in \ilaje. Besides the contrastive consonant  nature, the oral and nasal vowels are also both contrastive in \ife and \ilaje dialects respectively.
Further analsyses of \ourdata{} reveal that the low nasalised vowel /\textipa{\~a}/ mostly follows ``gh'' in \ilaje while the back lower- mid nasalised vowel /\textipa{\~O}/ accompanies ``gh'' in \ife dialect. One remarkable semantic variation is that standard \yoruba dialect uses ``s\d{o}'' and ``wi pe'' as \emph{say/talk},  however for \ilaje and \ijebu the morpheme mostly used is 
 ``f\d{o}'' while \ife uses ``ghii'', all of which have the same semantics.

\section{\ourdata{} Corpus}\label{sec:corpus_curation}
We curated parallel text and recorded high quality speech data across \ife, \ijebu, \ilaje, and Standard \yoruba dialects. Our data curation process involves three main steps: (i) text curation and dialect localization; (ii) speech recording; and (iii) text and audio alignment. 
\subsection{Text Curation and Dialect Localization}\label{sec:localization} We collected textual Standard \yoruba data from the following sources: (i) Bible study manuals;\footnote{\url{https://faithrebuilder.org/conference-bible-study-manuals}}
(ii) the \yoruba portion of MTTT,  a collection of multitarget bitexts based on TED Talks \cite{duh18multitarget}; and (iii) \yoruba news articles within the MAFT corpus \cite{alabi-etal-2022-adapting}. Given resource limitations and the demanding nature of this task, we gathered 352 sentences from the Bible study manuals, 247 sentences from TED Talks, and 907 sentences from news articles, amounting to a total of 1,506 sentences. Next, we proceeded to localising the compiled Standard \yoruba text into the three respective dialects: \ife, \ijebu, and \ilaje by recruiting trained linguists and translators who are literate and also native speakers of the respective dialects. We hired two translators or linguists per dialect and gave each a different domains to localise. The localisation process took about six to eight weeks and this included the localisation, quality assessment and incorporation of corrections. We provided monetary compensation for the localisation of the text. 
\begin{table*}[h]
\centering
\small
\begin{tabular}{l |rrrr|rrrr}
    \toprule
    
     & \multicolumn{4}{c}{\textbf{BLEU} $\uparrow$} 
     & \multicolumn{4}{c}{\textbf{AfriCOMET} $\uparrow$} \\
     \cmidrule(lr){2-5}\cmidrule(lr){6-9}
     
     & \ijebu & \ife & \ilaje  & Standard 
     & \ijebu & \ife & \ilaje  & Standard \\
     \midrule
     
    \textbf{M2M100} & 0.00 & 0.49 & 0.25  & 0.49
        & 0.26 & 0.27  & 0.26 & 0.30  \\
    \textbf{NLLB-600M} & 7.26 & 7.52 & 5.78  & 16.51
        & 0.52 & 0.50  & 0.49 & 0.65 \\
    \midrule
    \textbf{GMNMT} & \textbf{18.24} & \textbf{17.16} & \textbf{12.66}  & \textbf{43.46}
        & \textbf{0.59} & \textbf{0.57}  & \textbf{0.56} & \textbf{0.74} \\
    \midrule
    \textbf{Menyo} & 2.76 & 2.66 & 1.57  & 7.49
        & 0.44 & 0.40  & 0.40 & 0.52 \\
    \textbf{MT0} & 5.81  & 6.68  & 4.61   & 17.22 & 0.52 & 0.50  &  0.47 & 0.65 \\
    \textbf{Aya} &  7.18&  7.71  & 4.91 & 16.46
        & 0.49 & 0.50  & 0.45  & 0.63 \\
    \bottomrule
\end{tabular}%
\centering
 \caption{Zero-shot MT evaluation across all models. Google Translate outperforms other systems and is more robust to dialectal variation. However, a significant performance gap remains compared to the Standard Yoruba dialect.
 }
 \label{tab:zero_shot_mt}
\end{table*}
\subsection{Speech Recording}
Speaker selection is crucial when creating an ASR corpus; a speaker should be fluent, literate, trained, and familiar with voice recording \cite{ogayo22interspeech, 47393}. Due to time constraints and speaker availability, we were only able to record speech in standard \yoruba, \ife, and \ilaje dialects,  leaving \ijebu for a later version of the dataset. We retained the linguists and translators who localised the standard \yoruba text into \ife and \ilaje dialects.  We then recruited two additional native speakers per dialect that are literate in rendering the localised text into audio. All dialectal voice talents received monetary compensation. We first conducted an interview
, then asked the new recruits to record random samples of the text and send the recordings for assessment. The audio and corresponding text are vetted, after which we selected native speakers with high reading competence, good voice texture, and reading pace. This brought the total number of voice talents per dialect to four. To ensure that each voice talent within a dialect recorded text across all domains, we divided text in each domain (religion, Ted, news) into four parts. Each person recorded roughly 375 sentences from each domain resulting in a total of 3 hours of speech per dialect.

Recording is conducted using the speech recorder application designed by the YorubaVoice project \cite{ogunremi2024iroyinspeech}. The text files were uploaded per domain for each speaker on the YorubaVoice Recorder app. We used an M1 Pro 2021 chip MacBook with an audio-technica AT2020USB-X microphone set-up in an anechoic and sound-isolated voice recording booth for the recording process. Each text is recorded at 48 kHz and the audio files are provided in 16 bit linear PCM RIFF format. The app generates metadata that includes a unique speaker ID, audio ID with corresponding text, and the audio file. Finally, all the recordings were subjected to a quality control process by the data coordinator. We manually verified that the correct text was aligned with the appropriate audio file and re-aligned them when necessary. We also discovered one empty audio file in a particular dialect and proceeded to delete it, along with its corresponding text-audio pairs in all other dialects.
\paragraph{Final data statistics }
In total, the text portion of \ourdata{} consists of 1506 parallel sentences per dialect and 6024 sentences overall, while the speech portion consists roughly 3 hours of audio each in standard \yoruba, \ife and \ilaje, resulting in 9 hours of speech in total. We split the text and audio pairs in each dialect into 804 training samples, 200 validation samples and 502 test samples.

\begin{table}[h!]
\centering
\resizebox{0.95\columnwidth}{!}{%
\begin{tabular}{cccc}
\toprule
 Dialect &  length (hours) & Avg. length (seconds)&  Avg. tokens \\
\midrule
 Standard & 2.93 & 6.99&  15.81 \\
 \ilaje & 3.30 & 7.89 &  15.84\\
 \ife & 3.03 & 7.23& 15.53 \\
  \ijebu & - & -& 15.25 \\

\bottomrule
\end{tabular}}
\caption{Statistics of \ourdata{}. The number of train, validation and test samples is consistently (804/200/502) for each dialect.}
\label{tab:data_stats}
\end{table}

\section{Zero-shot Experiments}\label{sec:zero_exps}
We start by evaluating the zero-shot performance of current state-of-the-art models on the test portion of \ourdata{}. Based on the results from this initial evaluation, we then adapt the top-performing zero-shot models by finetuning on the training portion of \ourdata{} and report results in \autoref{sec:finetuning_setup}. MT experiments are conducted on all dialects, while ASR and S2TT experiments are conducted on all expect \ijebu . 

\subsection{Machine Translation}
We evaluate two classes of translation systems: MT-specific models and LMs. Here, the MT-specific
models use an encoder-decoder architecture and are
trained on large amounts of parallel data in multiple languages, whereas
the LMs are decoder-only models trained to maximize likelihood (i.e., next-token prediction) on text in multiple languages. All models we evaluate have standard \yoruba text in their training data. We only evaluate translation from the standard language or dialect into English since these experiments are zero-shot and we cannot expect the models to generate text in one of the dialects. This essentially enables us to measure the robustness of all of these models to variation in the \yoruba language. 

\begin{table*}[hbtp]
\centering
\small
\begin{tabular}{l |rrr |rrr}
    \toprule
    
     & \multicolumn{3}{c}{\textbf{ASR (WER) $\downarrow$}} 
     & \multicolumn{3}{c}{\textbf{S2TT (BLEU) $\uparrow$}} \\
     \cmidrule(lr){2-4}\cmidrule(lr){5-7}    
     &  \ife & \ilaje  & Standard 
     & \ife & \ilaje  & Standard  \\
     \midrule
    \textbf{MMS}  & \textbf{85.38} & \textbf{83.79} &  \textbf{72.50}
        & -  & - & - \\
    \textbf{SeamlessM4T} & 96.14  & 101.99  & 80.14
        & 5.52 & 3.30  & 13.16  \\      
    \textbf{Whisper} &  104.50  & 127.21 & 130.96
    &  0.17  & 0.21  & 	0.23  \\
    \bottomrule
\end{tabular}%
 \caption{Zero-shot performance on automatic speech recognition and speech translation.
 }
 \label{tab:asr_s2tt_zeroshot}
\end{table*}
\paragraph{MT-Specific Models} We evaluate M2M-100 \cite{fan2020englishcentric}, NLLB \cite{costa2022no}, and \menyo \cite{adelani-etal-2021-effect}. M2M-100 and NLLB are multilingual MT models trained on data spanning 100 and 202 languages respectively.
\menyo is a \yoruba-to-English-specific model fine-tuned on top of the multilingual pretrained mT5 model \cite{xue-etal-2021-mt5}. \menyo's model is trained with the \menyo dataset, a curated multi-domain standard \yoruba dataset with proper orthography.

\paragraph{Language Models} We evaluate two multilingual LMs, Aya \cite{ustun2024aya} and MT-0 \cite{muennighoff-etal-2023-crosslingual}, trained on 101 and 46 languages, respectively (standard \yoruba included). We prompt the LM to generate translations in a zero-shot setting with the prefix “Translate to English: " added to each sentence and greedily decode the continuation. We do not provide in-context examples in order to create a comparable setting to the evaluation of MT-specific models.

Finally, we include Google Translate (GMNMT)\footnote{\url{https://translate.google.com/}. API last accessed on June 7, 2024.} due to its widespread commercial use. We request the NMT model through the API, and cannot control any other aspects of its usage.

\paragraph{Results}
We measure translation quality using AfriCOMET \cite{wang2023afrimte} and BLEU \cite{papineni-etal-2002-bleu}. Firstly, we report zero-shot performance across all models in \autoref{tab:zero_shot_mt}.
Although performance is relatively low across the board, among MT-specific models, NLLB performs best across all dialects, outperforming M2M100 and \menyo. Comparing performance on LMs, Aya performs better than MT0 on all dialects except standard \yoruba. Google Translate outperforms all  systems across all dialects. Overall, we see a huge performance gap between standard \yoruba and the rest of the dialects. 
This observation is not surprising and is very consistent across all systems. The results in \autoref{tab:zero_shot_mt} also show that \ilaje has the worst-performing BLEU score across all models. We hypothesize that this is because \ilaje is largely spoken in \d{\`O}nd\d{\'o} state, which is geographically distant from \d{\`O}y\d{\'o} state where standard \yoruba originated from.

\subsection{Automatic Speech Recognition}
We evaluate three models: Whisper \cite{radford2022robust}, SeamlessM4T \cite{communication2023seamlessm4t}, and  MMS \cite{pratap2024scaling}. All models include standard \yoruba in their pretraining data. Whisper is an end-to-end ASR model, implemented as an encoder-decoder transformer, trained on 680,000 hours of multilingual and multitask supervised data collected from the web. The authors argue that it is robust to accents and variations in speech. It was optimized to perform the tasks of transcribing audio into its original language and translating the audio into English text. SeamlessM4T is a multilingual and multi-modal model that also translates and transcribes across speech and text. It is trained on 470,000 hours of mined speech and text-aligned data and supports ASR, S2TT, speech-to-speech translation, text-to-text translation and text-to-speech translation, although our focus here is ASR and S2TT. MMS is an ASR-only model finetuned on top of wav2vec 2.0 \cite{NEURIPS2020_92d1e1eb} models across 1,107 languages. In addition to dense finetuning, they also finetune language-specific adapter modules \cite{pmlr-v97-houlsby19a} for each language in their pretraining data.

\paragraph{Results}
We report word error rate (WER) with the models MMS, SeamlessM4T, and Whisper in \autoref{tab:asr_s2tt_zeroshot} (left).  Performance is generally poor across all models, with MMS performing the best. We hypothesize that MMS performs best due to its training with parameter-efficient finetuning using language-specific adapters. We see an average performance gap of 12 points between standard \yoruba and the other dialects on MMS and SeamlessM4T. With Whisper, the case is different: while the WER is generally very high, we see that only \ife is substantially better across all dialects. Upon manually reviewing the transcriptions from all models, we noticed that Whisper did not include diacritics in its generated transcriptions. \yoruba is a tonal language, and diacritics play a crucial role in disambiguating word meanings. We believe that this, coupled with the generation of overly segmented transcriptions contributes to Whisper's exceptionally high word-error rate exceeding 100.

\subsection{Speech Translation}
We only evaluate  Whisper \cite{radford2022robust} and SeamlessM4T \cite{communication2023seamlessm4t}. Just like MT, we only evaluate translation from the standard language or dialect into English as we cannot expect the models to generate text in any of the dialects without explictly finetuning it do so.

\paragraph{Results}
In \autoref{tab:asr_s2tt_zeroshot} (right), we present the zero-shot speech-to-text translation (S2TT) results of SeamlessM4T and Whisper models, the only open-source models we are aware of that include coverage for Standard \yoruba. Among all the tasks we evaluated, S2TT appears to be the most challenging. Performance is absolutely low for both models with Whisper performing particularly poorly. Across dialects, with SeamlessM4T, Standard Yoruba performs better yet again with an average of 9 points performance gap compared to \ilaje and \ife.

\section{Finetuning Experiments}\label{sec:finetune_exps}
\subsection{Machine Translation}\label{sec:finetuning_setup}
Next, we finetune NLLB-600M \cite{nllbteam2022language} on the training portion of our dataset in both directions, English$\rightarrow$Dialect and Dialect$\rightarrow$English. We experiment with training all dialects jointly under the \yoruba language code, and training the dialects separately by adding new language codes for each dialect and initializing them with the \yoruba embedding. In an attempt to further boost performance, we augment our training data with 10k instances from \menyo \cite{adelani-etal-2021-effect}.\footnote{\menyo was included in NLLB's pretraining data, however we try to include it in another step of language-specific finetuning.}

\paragraph{Results}
In \autoref{fig:zero_shot_GT_nllb} we analyze the translation quality following NLLB finetuning from Dialect$\rightarrow$English, comparing it with both the translation quality prior to finetuning and with Google Translate, which serves as the top-performing zero-shot system (\autoref{tab:zero_shot_mt}). Our results demonstrate that with only 802 training instances per dialects we outperform Google Translate on the non-standard dialects. While the performance of Google Translate remains notably superior for the standard dialect, we anticipate that scaling up the data could potentially bridge this gap. 
\begin{figure}[h]
    \centering
    \includegraphics[width=\linewidth]{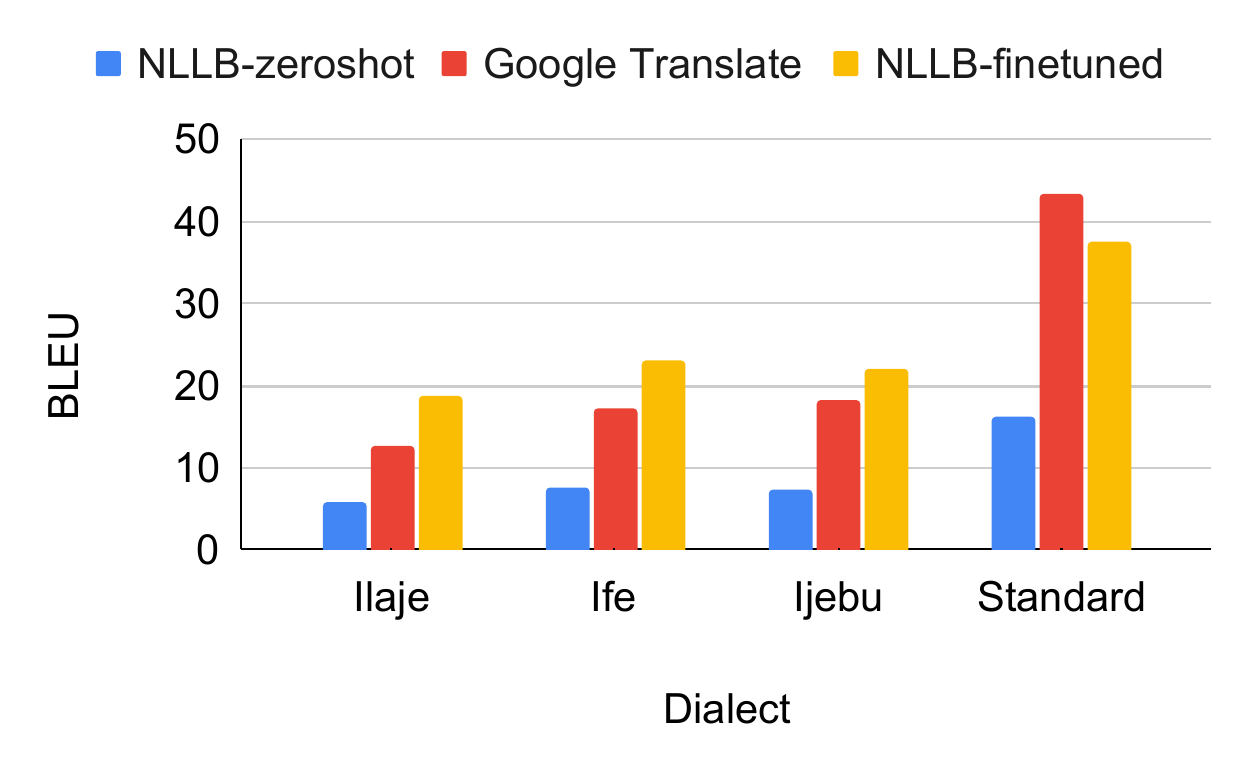}
    \caption{MT results $(\uparrow)$. We compare BLEU across Google Translate, NLLB prior to finetuning, and NLLB after finetuning.
    }
    \label{fig:zero_shot_GT_nllb}
\end{figure}

We present results for fine-tuning from English$\rightarrow$Dialect in \autoref{table:en2yo} in the Appendix. Our observation is that performance is generally worse than fine-tuning in  Dialect$\rightarrow$English direction. This is consistent with previous findings that translating into English 
could be easier than translating from it \cite{belinkov-etal-2017-neural}.

\subsection{Automatic Speech Recognition}
 We finetune MMS \cite{pratap2024scaling} and XLSR-Wav2Vec2 \cite{NEURIPS2020_92d1e1eb}. For the MMS model, we only finetune the \yoruba adapter layer, while the other weights of the model are kept frozen. 
\paragraph{Results}
We compare performance after finetuning XLSR and MMS  with two different model sizes each: 300M and 1.3B parameters. MMS is a more suitable choice for finetuning because of its parameter efficiency, since we only have to tune the \yoruba adapter layers. However, we choose to compare it with XLSR as well, as previous studies have reported significant performance improvements by finetuning XLSR \cite{ogunremi2024iroyinspeech}. In \autoref{fig:all_asr_finetuned}, we first see that for XLSR, fine tuning a model with less capacity (300M parameters) yields better performance across all dialects compared to fine tuning a model with about $4\times$ more parameters. However, with MMS, we see that finetuning the 1.3B model yields a lower WER compared to finetuning the 300M model. Here, the performance gap is not as drastic as with XLSR.  
\begin{figure}[h]
    \centering
    \includegraphics[width=\linewidth]{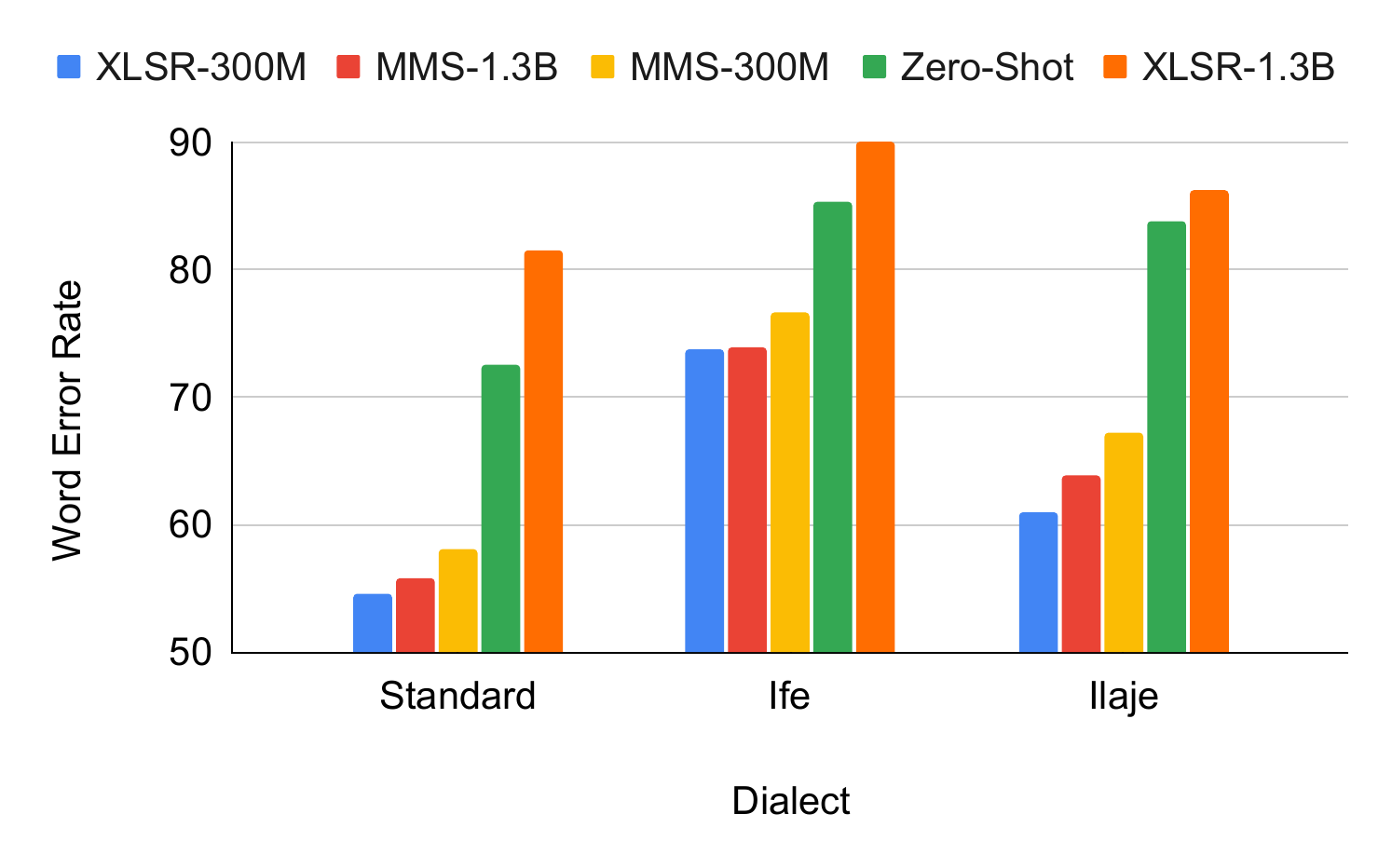}
    \caption{ASR results. $(\downarrow)$ We compare WER between zero-shot and jointly fine-tuning on all dialects on XLSR and MMS models.  
    }
   \label{fig:all_asr_finetuned}
\end{figure}

On average, there is a performance improvement of approximately 20\% after finetuning. As expected, across all models, the performance on the Standard \yoruba dialect remains considerably better than that of \ilaje  and \ife. We expect that increasing the size of the finetuning data could help close this gap and could be addressed in future work.

\subsection{Speech-to-Text Translation}
SeamlessM4T \cite{communication2023seamlessm4t} is the only model we finetune for speech-to-text-translation, since it its the best performing model from zero-shot experiments (see \autoref{tab:asr_s2tt_zeroshot} and the only other S2TT model (to the best of our knowledge) with Yoruba in its training data asides. We finetune in the  (Dialect$\rightarrow$English) direction. 

\paragraph{Results}
\begin{figure}[h!]
    \centering
    \includegraphics[width=\linewidth]{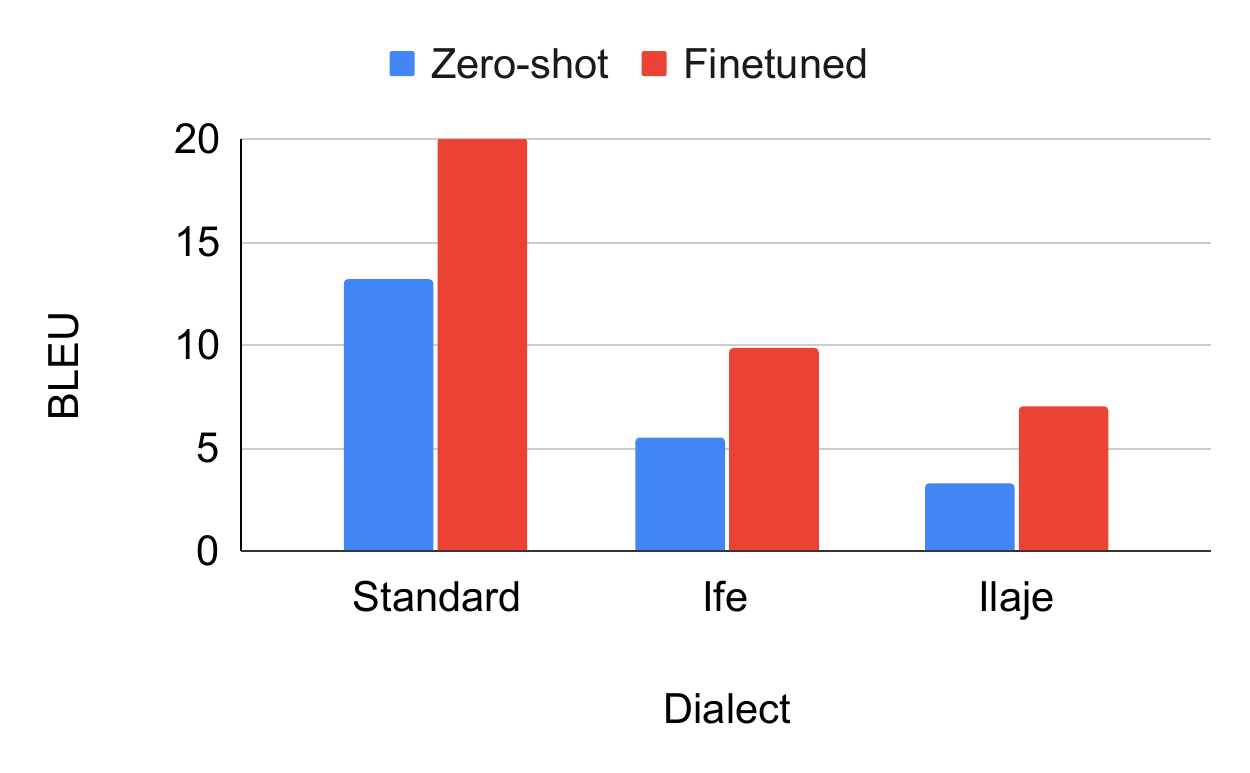}
    \caption{S2TT results $(\uparrow)$. We compare BLEU prior to finetuning and after finetuning SeamlessM4T.}
    \label{fig:s2tt_finetuned}
\end{figure}

The results in \autoref{fig:s2tt_finetuned} show that while we can reasonably boost performance on Standard \yoruba after finetuning, it still remains a hard task for the other dialects with just finetuning. We hypothesize that this occurs for two reasons, firstly the amount of \yoruba S2TT data in SeamlessM4T is smaller than the data available to train ASR \cite{communication2023seamlessm4t}. Secondly, while there is notable lexical variation across \yoruba dialects, the differences are even more pronounced in spoken language. This significant variation in pronunciation and intonation, coupled with the fact that S2TT data for \yoruba is scarcer than ASR data makes the task of adaptation particularly challenging.

\section{Human Evaluation}
We complement automatic evaluation metrics with a human evaluation study to assess translation and transcription quality from the best models after fine-tuning for MT and ASR. Previous research  has shown that word error rate (WER) is not nuanced, as it treats all errors in ASR text—insertions, deletions, and substitutions—the same, without considering their impact on readability \cite{Itoh2015AMF}.\footnote{\url{https://machinelearning.apple.com/research/humanizing-wer}} For ASR, one native speaker per dialect rated the quality of 30 randomly sampled transcriptions from the test set produced by our best ASR models after finetuning. After listening to the source speech they assess fluency (how natural and grammatically correct the transcription sounds in their dialect) and adequacy (how accurately the transcription conveys the meaning of the source speech) using a Likert scale of (1--5), the higher the better. In \autoref{tab:human_eval_asr} we show that human raters consider the transcriptions of standard \yoruba and \ife to be moderately adequate and fluent on average, compared to \ilaje. These findings align with our observations from automatic metrics.

\begin{table}[h!]
\centering
\resizebox{0.75\columnwidth}{!}{%
\begin{tabular}{crr}
\toprule
 & Adequacy $\uparrow$ & Fluency $\uparrow$ \\
\midrule
 Standard & 3.37 & 3.03 \\
 \ilaje & 2.73 & 2.62 \\
 
 \ife & 3.40 & 2.90  \\

\bottomrule
\end{tabular}}
\caption{Average human ratings of adequacy and fluency of transcriptions from the best ASR models after finetuning.}
\label{tab:human_eval_asr}
\end{table}

For MT, we ask human raters to compare the quality of translations from Google Translate with translations after finetuning NLLB, still focusing on fluency and adequacy still using a Likert scale (1--5). We provide the exact phrasings of instruction in the \autoref{sec:human_eval_instruct}. Our results, displayed in \autoref{tab:human_eval_mt}, show that Google Translate is rated to be more fluent and accurate on Standard \yoruba and \ilaje. However, our finetuned NLLB-600M model is rated to be more more fluent and accurate on \ife and \ijebu. The results on standard  \yoruba, \ife and \ijebu are very consistent with automatic evaluation results in  \autoref{fig:zero_shot_GT_nllb}. This is not the case with \ilaje, as our ratings are lower compared to Google Translate, which contrasts with our automatic evaluation in \autoref{fig:zero_shot_GT_nllb}.

\begin{table}[h!]
\centering
\resizebox{0.98\columnwidth}{!}{%
\begin{tabular}{crrrr}
\toprule
 & \multicolumn{2}{c}{Adequacy $\uparrow$} & \multicolumn{2}{c}{Fluency $\uparrow$} \\
\cmidrule(lr){2-3} \cmidrule(lr){4-5}
 & GMNMT & NLLB & GMNMT & NLLB \\
\midrule
Standard & 4.47 & 4.13 & 4.73 & 4.60 \\
\ilaje & 2.73  & 2.63 & 2.10 & 1.83 \\
\ife & 2.90 & 3.67 & 2.73 & 3.57 \\
\ijebu & 3.37 & 3.96 & 3.60 & 4.20 \\
\bottomrule
\end{tabular}}
\caption{Average human ratings of adequacy and fluency of test set translations comparing Google Translate with the best models after fine-tuning NLLB-600M}
\label{tab:human_eval_mt}
\end{table}

\section{Analysis and Discussion}
\paragraph{Does edit distance explain performance gaps?}
In this analysis we aim to understand how dialectal similarity influences model adaptation during finetuning. Ideally, we expect dialects with higher similarity to Standard \yoruba to perform better. Edit distance \cite{levenshtein1966binary} is a simple method commonly used in dialectometry to infer pronunciation differences between language dialects \cite{Nerbonne20205DF, Nerbonne1996PhoneticDB, Heeringa2004MeasuringDP}. In our work, we use edit distance as a proxy for similarity between Standard \yoruba and the other dialects in our corpus, expecting that dialects with a higher degree of similarity (lower edit distance) will perform better. We compute the average edit distance per dialect,
$\bar{d} = \frac{1}{N} \sum_{i=1}^{N} d(s_i, t_i)$, where \( N \) is the number of sentences in the test set of the dialect, \( s \) is the sentence in Standard \yoruba, \( t \) is the sentence in the corresponding dialect, and \( d(s_i, t_i) \) is the edit distance between \( s_i \) and \( t_i \) at the character-level.

We present the results of this analysis in MT in \autoref{tab:edit_dist_mt}. As expected, \ife has the smallest edit distance from Standard \yoruba and respectively also the best performance after finetuning. However we surprisingly see that while \ijebu has a higher edit distance than \ilaje, the model performance is higher for \ijebu. We conclude that edit distance has a weak correlation with our MT metrics.

\begin{table}[h!]
\centering
\resizebox{0.95\columnwidth}{!}{%
\begin{tabular}{cccc}
\toprule
\textbf{Dialect} & \textbf{Avg. ED} & \textbf{BLEU} & \textbf{AfriCOMET}  \\
\midrule
 \ife & 24.66 & 22.97 & 0.59 \\
 
 \ilaje & 38.07 & 18.64 & 0.55 \\

 \ijebu & 41.46 & 21.98 &  0.60 \\

\bottomrule
\end{tabular}}
\caption{Average edit distance and MT-Metrics comparison for MT across dialects.}
\label{tab:edit_dist_mt}
\end{table}

For ASR, we compute edit distance on phonetic transcriptions using the PanPhon library developed by \cite{Mortensen-et-al:2016}. The phonetic edit distance between standard \yoruba to \ilaje and \ife is 34.99 and 44.4, respectively. Here again, we also see no correlations between edit distance and performance on dialect adaptation.

\paragraph{Joint vs.~dialect-specific finetuning.}
Dialects often exhibit rather subtle variations in text and speech. In data-constrained scenarios like ours, it is reasonable to expect that jointly finetuning on all dialects would result in better performance compared to fine-tuning on each dialect individually. In our earlier finetuning experiments detailed in \autoref{sec:finetune_exps}, we explored joint training. Now, we try to compare performance between joint training and individual training on MT and ASR tasks. We generally see that on both tasks, joint training is beneficial. In MT, \autoref{table:mt_finetuning} in the Appendix shows a huge drop in performance across all dialects when we finetune on each dialect individually. This suggests that by jointly finetuning, the model leverages shared features across dialects for mutual benefit. However, in ASR, as shown in \autoref{tab:asr_all_models}, the drop in performance with individual finetuning is not as pronounced as with MT. We believe that in this case, the subtle variations in speech are sometimes significant, making it more challenging to greatly benefit from joint training. We however acknowledge that the data size of each individual dialect is one-fourth of the whole training set, so data paucity might also be influencing these results.

\section{Conclusion }
We introduce \ourdata ---the first high quality parallel text and speech corpus for four \yoruba dialects sourced primarily from native speakers, to enable ASR, MT and S2TT tasks for widely-spoken varieties of \yoruba. We have provided a detailed documentation of data curation process from standard text creation, to dialect localization and speech recording in communities where these dialects are spoken. Extensive experiments reveal that current models are not robust to dialectal variation, and improve significantly after our dialect-adaptive finetuning. Overall, our data collection methodology, new resources and improved models take a  
step towards enhancing the quality and equity of NLP technologies for \yoruba dialects and potentially other African languages.





\section*{Ethical Considerations}
Our datasets and models will be publicly released under an open license to foster research and continue to promote the development of NLP tools for African languages. Transcriptions, recordings and translations are carried out by paid native speakers who provided consent to use their voice to train our models. We acknowledge that the limited size of the corpus might not represent perfectly communities and speakers of the dialects. Further,  
dialectal generations, particularly when erroneous, could be perceived as biased
or even microaggressions by some native speakers, as well as dialect-specific errors from the models \cite{wenzel2024designing}. While our work provides resources that aim to reduce dialectal biases and unfairness in multilingual NLP systems, future work should focus on careful human evaluation of how these resources are incorporated in end-user tools. 


\section*{Limitations}
A limitation of our work is the robustness of the metrics we use for evaluation. While all of these metrics are standard for all of the tasks, we acknowledge that model-based metrics like AfriCOMET \cite{wang-etal-2024-afrimte} could be biased towards standard dialects that their models have been trained on. Exploring model-based metrics that facilitate robust evaluations on dialectal tasks remains a challenge for future work \cite{faisal2024dialectbench}. 

Additionally, the text portion of our dataset is translated from the standard dialect into English and the non-standard dialects. We acknowledge that this could introduce translation artifacts known as translationese \cite{Volansky2015OnTF} that are not present in the source dialect. However, we believe that the benefits of our dataset outweighs the potential risks of these artifacts.


\clearpage
\bibliography{anthology,custom}

\clearpage
\appendix
\section{Appendix}\label{sec:appendix}
 \subsection{Related Work}
 Previous works that have developing technologies and resources for machine translation \cite{ahia-etal-2021-low-resource, adebara-etal-2022-linguistically, adebara2021translating, lee-etal-2023-last, akinade-etal-2023-varepsilon, adelani-etal-2021-effect}, automatic speech recognition \cite{ogunremi2024iroyinspeech, communication2023seamlessm4t, NEURIPS2020_92d1e1eb} and speech translation \cite{communication2023seamlessm4t, oneata2024translating}  for \yoruba  have largely focused on the standard \yoruba dialect. This is because, just like other African languages, standard \yoruba is also very low-resourced, and all efforts have been directed there. Several works have shown that models often exhibit performance disparities  between standard languages and their dialectal counterparts \cite{diab-2016-processing, nigmatulina-etal-2020-asr, kantharuban-etal-2023-quantifying, ziems-etal-2023-multi, faisal2024dialectbench, ahmadi2024language, Joshi2024NaturalLP, blaschke-etal-2023-survey, aji-etal-2022-one, abdul-mageed-etal-2023-nadi}. Arabic language has roughly 30 regional dialects. Whilst majority of work has being done on Modern Standard Arabic (MSA), Arabic still  has the widest coverage of tasks and datasets across several of its dialects \cite{faisal2024dialectbench, diab-habash-2012-arabic, bouamor-etal-2018-madar, kchaou-etal-2020-parallel}. Within African languages, some works that aim to build dialect-aware models have conducted their studies on Igbo \cite{emezue2024igboapi}, Luhya \cite{siminyu-etal-21-luhya, chimoto-bassett-2022-low}, Bemba \cite{sikasote-anastasopoulos-2022-bembaspeech} and Kiswahili \cite{siminyu-etal-2022-corpus}.  

\subsection{Finetuning setup}
For the MT, we fine-tuned in both directions with a learning-rate of 2e-5 and batch size of 16. We trained for four epochs, and kept the model with the best eval loss. We used a weight decay of 0.01, warmup ratio 0.1, and a cosine annealing scheduler for learning rate. While for ASR finetuning, we fine-tuned with a learning-rate of 1e-3 and batch size of 8 for 20 epochs, as the validation WER continued to drop after preliminary runs with 10 epochs. For S2TT, we fine-tuned for 10 epochs with an optimal learning rate of 3e-4. All training was done on  two NVIDIA A40 GPUs.

\begin{table}[h]
    \centering
    \begin{tabular}{cccc}
        \toprule 
       Model & Standard & Ife & Ilaje  \\
        \midrule
        Zero-Shot & 72.50 & 85.38 & 83.79 \\
        MMS-300m-Individual & 74.67 & 93.20 & 78.24 \\
        MMS-1.3bn-Individual & 55.43 & 72.00 & 61.80 \\
        XLSR-300m-Individual & 56.26 & 81.23 & 64.22 \\
        XLSR-1.3bn-Individual & 67.65 & 78.70 & 76.36 \\
        MMS-300m-Joint & 58.11 & 76.58 & 67.17 \\
        MMS-1.3bn-Joint & 55.73 & 73.95 & 63.94 \\
        XLSR-300m-Joint & 54.55 & 73.72 & 61.03 \\
        XLSR-1.3bn-Joint & 81.57 & 90.04 & 86.30 \\
       
        \bottomrule
    \end{tabular}
    \caption{ASR Performance of across all models after fine-tuning individually and jointly}
    \label{tab:asr_all_models}
\end{table}

\subsection{Results from Joint vs Individual MT fine-tuning}
We present tables comparing jointly fine-tuning to individual fine-tuning on MT across the two training directions in \autoref{table:en2yo} and \autoref{table:mt_finetuning}.

\subsection{Human evaluation}\label{sec:human_eval_instruct}
We provide exact instructions given to human evaluators for our ASR and MT tasks in \autoref{tab:human_eval_asr} and \autoref{tab:human_eval_mt}
 
 \begin{table*}[t]
    \centering
    \begin{tabularx}{\textwidth}{lX}
        \toprule
        \multirow{10}{*}{\rotatebox[origin=c]{90}{\parbox[c]{9cm}{\large\textbf{}}}} & 
        You are tasked to evaluate the performance of an Automatic Speech Recognition (ASR) system on your native Yoruba dialect. This task involves assessing the accuracy and quality of transcriptions produced by this system when transcribing audio from a folder that will be provided to you. Your evaluations will help us understand how well these systems handle linguistic variations. Each filename has a corresponding audio file with the same name in the audio folder. Listen to the audio first, then look at the transcription from the model. Next, evaluate the quality of the transcription compared to the audio you listened to and provide a score in the Excel sheet. \\
        \midrule
        \textbf{Fluency} & Evaluate how natural and grammatically correct the transcription sounds in your dialect. \\
        & 1. \textbf{Incomprehensible}: The transcription is completely unintelligible and nonsensical. The text is difficult to understand. \\
        & 2. \textbf{Poor grammar and disfluent}: The transcription contains significant errors in grammar, syntax, and vocabulary that affect the clarity and naturalness of the text. \\
        & 3. \textbf{Grammatically correct, potentially unnatural}: The transcription is grammatically correct but may have some errors in spelling, word choice, or syntax. \\
        & 4. \textbf{Fluent and natural}: The transcription contains no grammatical errors, and the text is somewhat easy to read and understand. \\
        & 5. \textbf{Perfectly fluent and natural}: The transcription is completely natural, grammatically flawless, reading as if written by a native speaker. \\
        \midrule
        \textbf{Adequacy} & Assess how accurately the transcription conveys the meaning of the source speech. \\
        & 1. \textbf{Nonsense/No meaning preserved}: All information is lost between the transcription and the source. \\
        & 2. \textbf{Very poor meaning preservation}: The transcription preserves little meaning from the source. \\
        & 3. \textbf{Moderate meaning preservation}: The transcription retains some meaning but still misses important details. \\
        & 4. \textbf{Good meaning preservation}: The transcription retains most of the meaning of the source. \\
        & 5. \textbf{Perfect meaning preservation}: The meaning of the transcription is completely consistent with the source. \\
        \bottomrule
    \end{tabularx}
    \caption{Automatic Speech Recognition (ASR) human evaluation guidelines}
    \label{tab:asr_human_eval}
\end{table*}

\begin{table*}[t]
    \centering
    \begin{tabularx}{\textwidth}{lX}
        \toprule
        \multirow{1}{*}{\rotatebox[origin=c]{90}{\parbox[c]{7cm}{\large\textbf{}}}} & 
         You are tasked to evaluate the performance of two Machine Translation systems on your native Yoruba dialect. This task involves assessing the accuracy and quality of translations produced by these systems, when translating from your dialect into English. Your evaluations will help us understand how well these systems handle linguistic variations. \\
        \midrule
        \textbf{Fluency} & Evaluate how natural and grammatically correct the translation sounds in the target language. \\
        & 1. \textbf{Incomprehensible}: The translation is completely unintelligible and nonsensical. The text is difficult to understand. \\
        & 2. \textbf{Poor grammar and disfluent}: The translation contains significant errors in grammar, syntax, and vocabulary that affect the clarity and naturalness of the text. \\
        & 3. \textbf{Mostly grammatically correct, potentially unnatural}: The translation has few grammatical errors and also has some errors in spellings, word choice, or syntax. The language may not be natural. \\
        & 4. \textbf{Grammatically correct and natural}: The translation contains few grammatical errors, the vocabulary is precise, and the text is easy to read and understand. \\
        & 5. \textbf{Perfectly fluent and natural}: The translation is completely fluent, sounds natural and is grammatically correct. \\
        \midrule
        \textbf{Adequacy} & Assess how accurately the translation conveys the meaning of the source speech. \\
        & 1. \textbf{Nonsense/No meaning preserved}: All information is lost between the translation and the source. \\
        & 2. \textbf{Very poor meaning preservation}: The translation preserves little meaning from the source. \\
        & 3. \textbf{Moderate meaning preservation}: The translation retains some meaning but still misses important details. \\
        & 4. \textbf{Good meaning preservation}: The translation retains most of the meaning of the source. \\
        & 5. \textbf{Perfect meaning preservation}: The meaning of the translation is completely consistent with the source. \\
        \bottomrule
    \end{tabularx}
    \caption{Machine Translation (MT) Human evaluation guidelines}
    \label{tab:mt_human_eval}
\end{table*}

\begin{table}[h]
\centering
\resizebox{\textwidth}{!}{
\begin{tabular}{l |cccc|cccc}
    \toprule
    
     & \multicolumn{4}{c}{\textbf{BLEU} $\uparrow$} 
     & \multicolumn{4}{c}{\textbf{AfriCOMET} $\uparrow$} \\
     \cmidrule(lr){2-5}\cmidrule(lr){6-9}
     
     & \ijebu & \ife & \ilaje  & Standard 
     & \ijebu & \ife & \ilaje  & Standard \\
     \midrule
     
    \textbf{Individual} & 16.53 & 16.04 & 12.98 & 30.27
        & 0.57 & 0.56  & 0.52 & 0.69  \\
    \textbf{Joint} & 21.98 & 22.97 & 18.64 & 37.55
        & 0.60 & 0.59 & 0.55 & 0.71  \\
    \textbf{Joint + \menyo} & 19.80 & 20.77 & 17.21 & 31.75
        & 0.54 & 0.59 & 0.60 & 0.71  \\
    \bottomrule 
\end{tabular}}
 \caption{MT Finetuning Evaluation using NLLB-600M in the \yoruba to English direction, training the dialects as individual languages, jointly under \yoruba, and jointly along with \menyo data.}
 \label{table:mt_finetuning}
\end{table}

\begin{table}[h]
\centering
\resizebox{\textwidth}{!}{
\begin{tabular}{l |cccc|cccc}
    \toprule
    
     & \multicolumn{4}{c}{\textbf{BLEU} $\uparrow$} 
     & \multicolumn{4}{c}{\textbf{AfriCOMET} $\uparrow$} \\
     \cmidrule(lr){2-5}\cmidrule(lr){6-9}
     
     & \ijebu & \ife & \ilaje  & Standard 
     & \ijebu & \ife & \ilaje  & Standard \\
     \midrule
     
    \textbf{Individual} & 8.48 & 8.74 & 5.78 & 18.32
        & 0.52 & 0.50 & 0.47 & 0.66  \\
    \textbf{Joint} & 8.71 & 8.93 & 6.48 & 18.98
        & 0.52 & 0.50 & 0.47 & 0.66  \\
    \textbf{Joint + \menyo} & 7.23 & 7.25 & 5.29 & 17.24
        & 0.50 & 0.48 & 0.44 & 0.65  \\
    \bottomrule
\end{tabular}}
 \caption{MT Finetuning Evaluation using NLLB-600M in the English to \yoruba direction.}
 \label{table:en2yo}
\end{table}

\end{document}